\documentclass[
]{ceurart}

\usepackage{tabularx}
\usepackage{booktabs}
\usepackage{caption}
\usepackage{geometry}
\usepackage{url}
\usepackage{dirtytalk}
\usepackage{graphics}
\usepackage{adjustbox}
\begin{document}
\copyrightyear{2021}
\copyrightclause{Copyright for this paper by its authors.
  Use permitted under Creative Commons License Attribution 4.0
  International (CC BY 4.0).}
\conference{FIRE 2021: Forum for Information Retrieval Evaluation, December 16–20, 2021, India}

\title{Overview of Abusive and Threatening Language Detection in Urdu at FIRE 2021}

\author[1]{Maaz  Amjad}[%
email=maazamjad@phystech.edu,
url=https://nlp.cic.ipn.mx/maazamjad/,
]
\address[1]{Instituto Politécnico Nacional (IPN), Center for Computing Research (CIC),  Mexico }

\author[2]{Alisa Zhila}[%
email=alisa.zhila@ronininstitute.org,
]
\address[2]{Ronin Institute for Independent Scholarship, United States}

\author[1]{Grigori Sidorov}[%
email=sidorov@cic.ipn.mx,
]

\author[3]{Andrey Labunets}[%
email=isciurus@gmail.com,
]
\address[3]{Independent Researcher, United States}

\author[1]{Sabur Butt}[%
email=sabur@nlp.cic.ipn.mx,
]

\author[4]{Hamza Imam Amjad}[%
email=hamzaimamamjad@phystech.edu,
]
\address[4]{Moscow Institute of Physics and Technology, Russia}

\author[1]{Oxana Vitman}[%
email=oksana.vittmann@gmail.com,
]

\author[1]{Alexander Gelbukh}[%
email=gelbukh@gelbukh.com,
]

\begin{abstract}
With the growth of social media platform influence, the effect of their misuse becomes more and more impactful. The importance of automatic detection of threatening and abusive language can not be overestimated. However, most of the existing studies and state-of-the-art methods focus on English as the target language, with limited work on low- and medium-resource languages. In this paper, we present two shared tasks of abusive and threatening language detection for the Urdu language that has more than 170 million speakers worldwide. Both are posed as binary classification tasks where participating systems are required to classify tweets in Urdu into two classes, namely: (i) Abusive and Non-Abusive for the first task, (ii)  Threatening and Non-Threatening for the second. We present two manually annotated datasets containing tweets labeled as:  (i) Abusive and Non-Abusive, (ii) Threatening and Non-Threatening. The abusive dataset contains 2400 annotated tweets in the train part and 1100 annotated tweets in the test part. The threatening dataset contains 6000 annotated tweets in the train part and 3950 annotated tweets in the test part. We also provide logistic regression and BERT-based baseline classifiers for both tasks. In this shared task, 21 teams from six countries registered for participation  (India, Pakistan, China, Malaysia, United Arab Emirates, Taiwan), 10 teams submitted their runs for Subtask A —Abusive Language Detection, 9 teams submitted their runs for Subtask B —Threatening Language detection, and seven teams submitted their technical reports. The best performing system achieved an F1-score value of 0.880 for Subtask A  and 0.545 for Subtask B. For both subtasks, m-Bert based transformer model showed the best performance.

\end{abstract}


\maketitle

\section{Introduction}

In cyberspace, abusive and threatening content is a glaring problem that has been present since the beginning of human interaction on the internet and will continue to persist in future. Social media platforms today are the venues for free expression for all communities, and community backlashes can result in a lot of negative externalities. Thus, with the growth of social media platforms and their audiences, regulating threatening and abusive content becomes a concern for the welfare of all stakeholders. Though leading platforms such as Twitter and Facebook have set up community standards for the prevention of cybercrimes, early detection of such content is vital for the safety of cyberspace. 

Detection of abusive and threatening text is a complex problem as the platforms find it challenging to maintain a balance between limiting the abuse and giving users ample freedom to express themselves. Failing to meet the balance can result in users losing trust in the platform as well as disengagement with the content. Platforms also find it challenging to detect texts in multiple languages, especially low resourced languages and code mixed languages. Manual filtering and understanding of this content is logistically daunting and resource unfriendly. It can also result in the delay of necessary and timely action needed in case of active threats and abuse. Hence, Natural Language Processing (NLP) researchers have been working on the early detection of threats and abuse by providing various automated solutions based on machine learning and deep learning in particular.

Several studies previously have attempted to deal with the problem of abusive language~\cite{naseem2019abusive, mubarak2017abusive} and threat detection~\cite{8934609, ashraf2020individual}. These problems have been attempted through supervised machine learning~\cite{waseem2016hateful, chen2012detecting, van2015detection} and deep learning~\cite{pavlopoulos2017deep, cecillon2021graph, wulczyn2017ex} approaches breaking it down into binary, multi-label, or multi-class classification problems. However, these attempts are only limited to European languages, Arabic, and a few South Asian languages such as Hindi, Bengali, and Indonesian.

Here we present a new shared task for abusive and threatening language detection in tweets written in Urdu. The task is aimed at driving attention and effort of the research community to developing more efficient methods and approaches for this vastly spoken language and highlighting difficulties that specific to the writing system and the use of Urdu.    
The paper describes the abusive and threatening language tracks~\footnote{\url{https://www.urduthreat2021.cicling.org}} organized by the authors 
within the Hate Speech and Offensive Content Identification (HASOC) evaluation tracks of the 13\textsuperscript{th} meeting of Forum for Information Retrieval Evaluation (FIRE) 2021~\footnote{\url{http://fire.irsi.res.in/fire/2021/hasoc}} 
and co-hosted by Open Data Science (ODS) Summer of Code initiative 2021~\footnote{\url{https://ods.ai/competitions}}. The task is comprised of two sub-tasks:

\begin{enumerate}
    \item Sub-task A: \textbf{Abusive language detection}~\footnote{\url{https://ods.ai/competitions/urdu-hack-soc2021}}. The task offered a dataset of Twitter posts ("tweets") in Urdu language split into the training part with the annotations available to participants and the testing part provided without annotations. 
     The dataset annotation procedure followed Twitter's definition of abusive language~\footnote{\url{https://help.twitter.com/en/rules-and-policies/abusive-behavior}} to identify posts that are abusive towards a community, group, or an individual as the ones meant to harass, intimidate or silence someone else’s voice. The tweets were annotated in a binary manner, i.e., \textit{abusive} or \textit{non-abusive}. 
     The participants were asked to determine the correct labels for the testing part and submit their annotations. The solutions were evaluated using F1 and ROC-AUC metrics. 
     
    \item Sub-task B: \textbf{Threatening language detection}~\footnote{\url{https://ods.ai/competitions/urdu-hack-soc2021-threat}}. Similarly, the task offered a dataset of tweets in Urdu annotated as \textit{threatening} or \textit{non-threatening} split into the training and testing parts, with the annotations of the testing part hidden from the participants.   
    The annotation procedure for the Sub-task B dataset followed Twitter's definition of threatening tweets~\footnote{\url{https://help.twitter.com/en/rules-and-policies/violent-threats-glorification}} as those that are against an individual or group meant to threaten with violent acts, to kill, inflict serious physical harm, to intimidate, or to use violent language. 
    The task and the evaluation procedure were identical to Sub-task A
    
\end{enumerate}

With these shared tasks our contributions are:

\begin{itemize}
    \item spreading awareness and motivating the community to propose more efficient methods for automated detection of abusive and threatening messages in social media in Urdu as well as providing means for standardized comparison as emphasized in Section~\ref{sec:importance};  
    \item collection and annotation of the largest so far datasets for abuse and threat detection in the Urdu language described in Section~\ref{sec:dataset} , in particular, 3500 tweets annotated as abusive or not and 9,9500 tweets annotated as threatening or not; 
    \item the train and test split that allows for a fair result comparison (see Section~\ref{sec:train-test} for details and grounds) not only among the current participants but also for future research; 
    \item provision of highly competitive baseline classifiers in Section~\ref{sec:baselines}; 
    \item overview and comparison of the submitted solutions for abusive language and threat detection in Urdu in Sections~\ref{sec:overview} and~\ref{sec:results}.
\end{itemize}

\section{Importance of Identifying Abuse and Threat in Urdu}\label{sec:importance}

Urdu is one of the largest spoken languages in South Asia. It is the national language of Pakistan and has its roots in Persian and Arabic bearing additional structural similarities with many languages from other language families, e.g., Hindi~\cite{ahmed2010developing, visweswariah2010urdu}. Urdu is spoken by more than 170 million~\footnote{\url{https://www.ethnologue.com/language/urd}} people worldwide and the number is increasing every day. Yet it lacks solutions and resources for the most essential natural language processing problems. 

Urdu is mostly written using the Nastal\'{i}q script. However, certain populations also use the Devanagari script which is normally used for writing Hindi. Hence, Urdu texts experience the phenomenon of \textit{digraphia} that is the use of more than one writing system for the same language. Additionally, Urdu is quite complex linguistically~\cite{adeeba2011experiences} as its morphological and syntactic structure is a combination of Turkish, Arabic, Persian, Sanskrit, and English. Hence, contributing to Urdu is also fundamental for the success of other languages. 

Population of the Urdu speaking countries have substantial access to social media, and millions of speakers are exposed to unregulated or poorly regulated hate, abuse, and threats. Various extremist and terrorist groups have developed communities on social media platforms that spread abuse, threats, and terror~\cite{bertram2016terrorism}. As they post in local languages, in particular, in Urdu, much of the content is left unchecked until reviewed and reported manually. Pakistan suffered decades of terrorism and had to resort to banning social media on several occasions to tackle terrorism~\cite{hassan2018social}. Hence, the development of resources in Urdu for threat and abuse detection is an urgent requirement for the safety of millions.

\section{Literature Review}

Offensive content encompasses a variety of phenomena including aggression~\cite{aroyehun2018aggression, farhan2019human}, sexism~\cite{butt2021sexism}, hate speech~\cite{djuric2015hate, waseem2016hateful}, threat detection~\cite{ashraf2020individual, 9536729}, toxic comment detection~\cite{obadimu2019identifying}, abusive language detection~\cite{naseem2019abusive, mubarak2017abusive} and many others. Previous research~\cite{waseem2017understanding} have attempted to distinguish various types of abuse such as implicit vs. explicit abuse or identity vs. person-directed abuse to identify more nuanced expressions of abuse. 

Multiple annotated datasets are available for a variety of offensive content phenomena sourced from numerous social media platforms and portals. Yahoo Finance corpus~\cite{djuric2015hate} comprises English language texts from the Yahoo's finance portal which is annotated into two classes: clean and hate speech. 
Research ~\cite{waseem2016hateful} collected a dataset of Twitter posts in English and annotated them into three classes as sexism, racism, or neither. 
Similarly, work ~\cite{davidson2017automated} also annotated tweets in English into three classes, yet different from work~\cite{waseem2016hateful}: hate speech, offensive language, and neither. 
On the contrast, study~\cite{founta2018large} distinguished four offensive classes in their collection of Twitter posts in English: hateful, spam, abusive, and neutral. 

Youtube has been another source for data collection for abusive language in English~\cite{obadimu2019identifying, threat2019} as well as in other languages, in particular, Arabic~\cite{mubarak2017abusive}. In particular, the study by Ashraf et al.~\cite{ashraf2020individual} is based on the YouTube comment and replies collection introduced by Hammer et al.~\cite{threat2019} with additional annotation of its subset as whether a threat is directed towards a group or an individual. Another study~\cite{ashraf_abusive_2021} collects a dataset of 2,304 YouTube comments with 6,139 replies in English and annotates it in two ways: a binary annotation for abusive language as well as a three class annotation for topics: politics, religion, and other.  

Attempts have been made to create threat and abuse detection models for Bengali language~\cite{8934609}. Posts and comments have been collected from different pages of Facebook.
Threatening and abusive language was labeled as ``YES'' in the dataset and the rest of the data which is not abusive was labeled as ``NO''.
For more detailed analysis of the available datasets, we recommend these studies~\cite {vidgen2020directions, naseem2019abusive, nakov2021detecting}.  

Apart from the papers proposing a single solution, a number of shared tasks have been organized to incentivize creation of multiple robust systems for offensive phenomena detection in texts. Some of the popular shared tasks are OffensEval~\cite{zampieri2020semeval, zampieri2019semeval} with available datasets in Greek, English, Danish, Arabic, Turkish, and English; GermEval 2018~\cite{wiegand2018overview} for texts in German; TRAC shared task~\cite{fortuna2018merging} for Hindi, English, and Bengali; SemEval-2019~\cite{basile2019semeval} for hate speech detection in English and Spanish; HASOC 2019 and 2020~\cite{mandl2019overview, mandl2020overview} for German, English, Tamil, Malayalam, and Hindi. 

Among the common approaches for offensive language detection, we observe feature-based approaches with traditional ML classifiers. Works ~\cite{chen2012detecting, van2015detection, waseem2016hateful, yin2009detection, mandl2019overview, mandl2020overview, basile2019semeval, zampieri2019semeval, zampieri2020semeval} use various combinations of features such as N-grams, Bag-of-Words (BOW), Part-of-Speech (POS) tags, Term Frequency—Inverse Dense Frequency (TF-IDF) representation, word2vec representation, sentiments, and dependency parsing features provided as input to the traditional ML models such as Support Vector Machines (SVM), Logistic Regression (LR), Random Forest (RF), Decision Tree (DT), Naive Bayes (NB), etc. 

Among the more efficient approaches for the task, we see boosting-based ensembles as well as neural networks, in particular, deep NNs such as transformers. For example, Ashraf et al.~\cite{ashraf_abusive_2021, ashraf2020individual} used n-gram and pre-trained word embeddings in combination with traditional ML (LR, RF, SVM, NB, DT, VotingClassifier, and the boosting-based ensemble AdaBoost) as well as Neural Network based methods (MLP, 1D-CNN, LSTM,  and Bi-LSTM) 
for the abusive language detection and for the prediction of an individual- vs. group-targeted threat correspondingly. While the BiLSTM approach achieved an F1 score of 85\%, the use of conversational context along with the linguistic features achieved even higher F1 score of 91.96\% using an ensemble AdaBoost classifier.

BiLSTM and Convolutional Neural Networks (CNN) were used to tackle abusive language and hate speech detection in multiple other works. Studies employing graph embeddings to learn graph representations from online texts~\cite{cecillon2021graph}, paragraph2vec~\cite{djuric2015hate}, and Recurrent Neural Networks (RNN) with attention~\cite{wulczyn2017ex}, RNN with Gated Recurrent Units (GRUs)~\cite{pavlopoulos2017deep} have also shown encouraging results. The pre-trained transformer methods such as RoBERTa, BERT, ALBERT and GPT-2 to detect hate speech detection can be seen achieving high accuracies~\cite{devlin2018bert, radford2019language, lan2019albert}. A recent study~\cite{vashistha2021online} applied XLM, BERT and BETO models to achieve promising results on similar tasks for hate speech detection. 

While each offensive subcategory uses different definitions for annotation, similar methods can be applied across the offensive content detection tasks. All these techniques can be used to test the best combinations for detection of abuse and threat in the Urdu language\cite{amjad2021threatening, amjad2021abusive} and our study opens vast avenues for researchers to achieve this goal.

\section{Datasets Collection and Annotation}
\label{sec:dataset}

\subsection{Threatening and Abusive Datasets Collection}
In the beginning, we created a dictionary of most used abusive and threatening words in Urdu. We used those words as keywords on Twitter to mine tweets containing more abusive and threatening words in Urdu, which we manually added to our dictionary. The dictionary includes words that appeared even ones to threat or abuse someone. This dictionary is publicly available for research purposes~\footnote{\url{https://github.com/MaazAmjad/Threatening_Dataset}}.Thus, we collected a sufficient number of abusive and threatening seed words which  were further used to crawl tweets through the Twitter Developer Application Programming Interface (API)~\footnote{\url{https://developer.twitter.com/en/docs/twitter-api/v1/tweets/search/api-reference/get-search-tweets}} using \textsc{Tweepy} library. 
Thus, we gathered enough words and phrases that are used to threat or abuse individuals. We collected tweets containing any of these keywords from our dictionary for a 20 month period from January 1st, 2018 to August 31st, 2019. At this time the general elections being held in Pakistan in July 2018. Typically, during the election season, people tend to be more expressive when supporting as well as opposing political parties. In total we crawled 55,600 number of tweets containing the
seed words. 

\subsection{Threatening and Abusive Datasets Pre-processing}
Since Urdu shared many common words in Persian, Turkish and Arabic, so when we crawled tweets using our initially collected words, the Twitter APA also crawled many non-Urdu tweets. Since this research was primary focused on Urdu lanaguage, we discarded all the non-Urdu tweets manually. Thus, two different datasets have been created:  (i) abusive dataset~\footnote{\url{https://github.com/MaazAmjad/Urdu-abusive-detection-FIRE2021}}, containing 3,500 tweets, 1,750 of them are abusive and 1,750 of them are non-abusive (ii) threatening dataset~\footnote{\url{https://github.com/MaazAmjad/Threatening_Dataset}}; containing 9,950 tweets, 1,782 threatening tweets and remaining tweets are non-threatening. 

\subsection{Threatening and Abusive Datasets Annotation}

We defined guidelines to annotate abusive and threatening tweets. 

To annotate the dataset the annotators have been recruited. All of them satisfied the following criteria:
(i) country of origin - Pakistan; (ii) native speakers of Urdu; (iii) are familiar with Twitter; (iv) aged 20–35 years;
(v) detached from any political party or organization;
(vi) have prior experience of annotating data;
(vii) educational level was a masters degree or above.

We computed Inter-Annotator Agreement (IAA) using
Cohen’s Kappa coefficient [39] as it is a statistic measure to
check the reliability between two annotators. 
We provided instructions with task definitions (which are reproduced below) and examples.
Hierarchical annotation schema was used and the main dataset was divided into two different datasets  to distinguish between
whether the language is threatening ot non-threatening, abusive or non-abusive. 
We followed Twitter definition to describe abusive~\footnote{\url{https://help.twitter.com/en/rules-and-policies/abusive-behavior}} and threatening~\footnote{\url{https://help.twitter.com/en/rules-and-policies/glorification-of-violence}} comments towards an individual or groups to harass, intimidate, or silence someone else’s voice.

\section{Training and Testing Dataset Split}
\label{sec:train-test}

Due to the requirements extended by the competition conditions and in purpose of fair evaluation of the participant's submission, a slightly larger portion of the datasets was withheld as corresponding testing parts than it would be done under `normal' data science operations. Namely, 40\% of the data was withheld for the Threatening Language task, and 32\%, for the Abusive Language task. This is done, first of all, to ensure that the testing set is non-trivial and represents well the variety of possible lexical expressions for both classes. Second, during the active period of the competition, the participants could observe the scores only from the ``public'' part of the testing set, whereas the scores on the ``private'' part of the testing set were made public only after the end of the competition. The partitioning of the test set into public and private is necessary to avoid pure guessing or tampering with predictions. We ensured that each partition of the testing data was large enough to compute a score that is sufficiently reflective of the actual performance of a system. The details are presented in Table ~\ref{tab:ttsplit}.

\begin{table}[!h]
  \caption{Descriptive table of the train/test split for abusive and threatening language datasets. In parentheses the fraction of positive, \textit{i.e.}, abusive or threatening respectively, labels is indicated. }
  { %
  \label{tab:ttsplit}
  \begin{tabular}{lcccc}
    \toprule
\textbf{Dataset} & \textbf{train}  & \textbf{test} & \textbf{test\textsubscript{PUB}} & \textbf{test\textsubscript{PRIV}}  \\
    \midrule
Abusive &  2400 \textit{(49.50\%)} & 1100 \textit{(48.81\%)} & 400  \textit{(48.50\%)} & 700 \textit{(47.29\%)}\\
Threat &  6000 \textit{(17.85\%)} & 3950 \textit{(18.20\%)} & 1000 \textit{(15.80\%)} & 2950 \textit{(19.02\%)}\\
  \bottomrule
\end{tabular}
}
\end{table}

To be clear, the participants were handed out the \textit{entire} test set without true labels. After a submission, the scores were shown only for the public partition of the test set. As it can be observed from Tables ~\ref{tab:overlapping2} and {\ref{tab:overlapping3}}, there still was some amount of shake-up among the scores and corresponding ranks on the public and private partitions.

Now that both the training and the testing sets along with their true labels are available to the research community, a different approach to train/test split may be possible. However, for a fair comparison with the competition submissions and results provided in this paper, we suggest following the original split.

\section{Evaluation Metrics}

The submitted systems were evaluated by comparing the labels predicted by the participants’ classifiers to the hidden ground truth annotations. For quantifying the classification performance, we computed the commonly used evaluation metrics: F1 score and ROC-AUC score. F1 score serves as a better metrics for unbalanced datasets than Accuracy and, therefore, accommodates our settings. The ROC-AUC score gives an estimate of the overall quality of the model at the various level of predicted confidence thresholds and serves as a more holistic evaluator.

\section{Baselines}
\label{sec:baselines}

For the competition, the organizers prepared three baseline systems: two of them reflected different aspects of traditional ML approach involving Bag-of-Words features and were meant to be lower boundary scoring baselines while the third system was based on the recent deep learning approach involving fine-tuning of the BERT model ~\cite{devlin2018bert}. 

\subsection{LogReg with Lexical Features}

All data pre-processing steps and most of the modeling details are identical for both subtasks, abusive and threat detection, if not explicitly indicated otherwise. 

First, all possible word unigrams and bigrams were extracted from the training dataset using the popular \textsc{NLTK}\footnote{https://www.nltk.org}~\cite{nltk2004} software package for NLP, v. 3.4.5, counting the numbers for n-gram occurrences in the dataset. Further, the occurrence threshold of 3 was applied to unigrams corresponding to the 75\textsuperscript{th}-percentile of all encountered unigrams. In other words, we took the top 25\% of most frequently occurring unigrams as features. Similarly, the 95\textsuperscript{th}-percentile occurrence threshold of {4} was applied to bigrams. We also added 2 additional features to account for Out-Of-Vocabulary (OOV) unigrams and bigrams correspondingly. Eventually, the feature set was comprised of the top occurring unigram features, top occurring bigram features, and the two OOV features. The statistics for each feature type by the subtask dataset and the total number of features is provided in Table~\ref{tab:feats1}.      

\begin{table}[!htb]
  \caption{Feature type statistics for each subtask for the first baseline solution.}
  { %
  \label{tab:feats1}
  \begin{tabular}{lcccc}
    \toprule
\textbf{Subtask} & \textbf{unigrams}  & \textbf{bigrams} & \textbf{OOV features} & \textbf{total features}  \\
    \midrule
Abusive &  1775 & 1212 & 2 & 2989\\
Threat &  3640 & 3413& 2 & 7055 \\
  \bottomrule
\end{tabular}
}
\end{table}

Further, each tweet instance was represented as a straightforward count of feature occurrences in the tweet, all OOV n-grams counting towards corresponding special OOV features. No normalization was done as all tweets have approximately the same length. 

Logistic regression was selected as the classifier algorithm for our traditional ML baseline solutions. In the first system, we used the implementation from \textsc{scikit-learn}~\footnote{\url{https://scikit-learn.org}~\cite{scikit-learn}} v. 0.22.1, which is a popular software package that includes a number of ML algorithms. The \texttt{max\_iter} parameter was set to 1000 to make sure the training converges. 

For the Threat Subtask dataset, where the positive and negative classes are imbalanced, we also set the \texttt{class\_weight} parameter to \textit{balanced} which ensured automatic instance reweighing. 

The code is available at the organizers's GitHub repository~\footnote{\url{https://github.com/UrduFake/urdutask2021/}}.

The balanced baseline secured the 8\textsuperscript{th} place on the Threat Subtask private leaderboard with F1-score equal to 0.49186, ROC-AUC, to 0.76991. The unbalanced version applied in the Abusive Subtask came 12\textsuperscript{th} on the private leaderboard scoring 0.78684 F1-score, 0.88295 ROC-AUC.

\subsubsection{A version of LogReg with lexical features and TF-IDF count}
\label{sec:logreg2}

We also submitted a variation of the Log-Reg based classifier with a few technical as well as conceptual modifications. Instead of a simple \textit{n}-gram occurrence count, the TF-IDF vectorization approach was used for text representation. For this, the \texttt{TfidfVectorizer} function from the \textsc{scikit-learn} package was used. It is to note that the types of features were unigrams only. The number of features was set as in the previous approach.

Another purely technical difference was that the LogReg classifier was implemented as a ``single node neural network'' which is algorithmically and equationally equivalent to logistic regression.

The implementation was done using the \textsc{PyTorch} framework~\footnote{\url{https://pytorch.org}} ~\cite{pytorch2019}. This training set-up converged much sooner, with mere 30 epochs,  or in terminology of traditional ML, iterations, for both datasets. The optimal number of epochs was determined using a validation dataset which was 10\% of the corresponding training data.

For the threatening language detection dataset, similarly to the previous approach, the dataset balancing was performed by applying \texttt{torch.nn.BCEWithLogitsLoss} function.

These differences in approaches were reflected in the final score difference. Interestingly, for the abusive language detection task, while this variant showed slightly higher scoring (0.77008 F1-score for this version vs  0.72928 F1 for the above version, and 0.86674 vs 0.85286 ROC-AUC) and, hence, the rank (11\textsuperscript{th} \textit{vs}. 13\textsuperscript{th}) on the public leaderboard, it actually showed same scores  on the private leaderboard, 0.78684 F1 and 0.88295 ROC-AUC, to the extent of decimal precision displayed, sharing the 12\textsuperscript{th} and 13\textsuperscript{th} ranks. 

More notably, in the threatening language detection task, the results and scores returned by the two versions, not only varied largely, but the score difference of the systems actually flipped significantly between the private and public leaderboards. On the public leaderboard, the \textsc{scikit-learn} version gained higher scores: 0.46471 F1 \textit{vs}. 0.45161 F1 for the \textsc{PyTorch} version, and 0.79502 ROC-AUC \textit{vs}. 0.78899 ROC-AUC for the \textsc{PyTorch} version. Yet on the private leaderboard the \textsc{scikit-learn} version gained less: 0.49186 F1 \textit{vs}. 0.51404 F1 for the \textsc{PyTorch} version, and	0.76991 ROC-AUC \textit{vs}. 0.78212 ROC-AUC, respectively. 

This brings us to a likely conclusion that,  no matter the ML package, for the abusive task, the LogReg classifier along with the lexical bag-of-word features is a sufficiently powerful tool that can properly converge on the provided dataset learning a coherent pattern. 

However, the threat detection task is a more complex task not only due to the label imbalance but also due to the intrinsic semantic complexity of the phrases, the latter having a much larger effect. Therefore, simple classifiers and purely lexical features are too weak to capture higher levels of semantic complexities and should not be relied on for this subtask.         

\subsection{BERT-based baseline}

The dataset sizes of 2400 and 6000 items along with training example length below 200 characters made the tasks approachable with transfer learning-based methods using foundational deep BERT-like models.

The proposed deep learning-based solutions for both subtasks, Abusive and Threat detection, used \textsc{pretrained multilingual uncased BERT}~\footnote{\url{https://huggingface.co/bert-base-multilingual-uncased}}~\cite{devlin2018bert} from huggingface transformers library ~\cite{wolf-etal-2020-transformers} as a base model.

Huggingface ``built-in'' \textsc{\texttt{BertForSequenceClassification}}~\footnote{\url{https://github.com/huggingface/transformers/blob/27d4639779d2d316a7c5f18d22f22d2565b84e5e/src/transformers/models/bert/modeling\_bert.py\#L1486}} class with 2 output units was selected as a classification head, 
where pooled output from [CLS] token is passed through a dropout layer, followed by a linear layer with output units leading into cross-entropy loss function.

For the Abusive Subtask, we split the provided training dataset into
TRAIN/DEV sets via a standard 80:20 ratio.
Using the TRAIN set, the model is further fine-tuned for the target classification task for 3 epochs with minibatch size of 32 and 60 minibatches per epoch.  The total number of minibatches, and correspondingly optimization steps, was 180.
The fine-tuning process used the DEV set to evaluate the model performance every 4 minibatches in order to load a model with the best F1 score from checkpoints at the end of the fine-tuning.

For the Threat Subtask, we split the provided training dataset into TRAIN/DEV sets via a 85:15 ratio. We deviated from the standard 80:20 split to let the model train with slightly more data and more negative examples as a result at the cost of less accurate F1 score.
Using the TRAIN set, the model was later fine-tuned for the target task 
for 5 epochs with minibatch size of 32 and 160 minibatches per epoch (total number of minibatches / optimization steps was 800). In our set-up, the model for the Threat Subtask converged slower than the one for the Abusive subtask, therefore we trained the network for 5 epochs instead of 3. The cross-entropy loss function additionally used inverse class sizes as weights to account for imbalance.
The fine-tuning used the DEV set to evaluate the model every 8 minibatches (not 4 due to longer training) in order to load a model with the best F1 score from checkpoints at the end of the fine-tuning.

The first baseline model for the Abusive Subtask came 3\textsuperscript{rd} on the private leaderboard with F1-score equal to 0.86221, ROC-AUC to 0.92194. The second baseline model for the Threatening Subtask came 9\textsuperscript{th} on the private leaderboard scoring 0.48567 F1-score, 0.70047 ROC-AUC. Considering the original BERT's ~\cite{devlin2018bert} scores at GLUE and other benchmarks, as well as further progress in language model pretraining ~\cite{liu2019roberta}, the first model's relatively high F1 score was expected. The Abusive Subtask was a sentence classification task with little specific constraints (such as overly large sequence length or similar obstacles), where deep bidirectional architecture-based and other large pretrained language models generally outperform traditional machine learning approaches in a number of domains. At the same time, better handling of class imbalance in the Threat Subtask could help the second baseline model achieve better convergence and a higher F1 score. We speculate, that domain-specific improvements at preprocessing, additional intermediate-task training, and complementary handcrafted features used along with the sentence embeddings can further boost the score for both models. In other words, subject matter knowledge of language and relevant threat landscape is indispensable for real-world threat and abuse detection in Urdu language. Finally, we see incorporating continued training and more domain-specific research in adversarial training, out-of-distribution detection, and outlier detection as viable directions to make a model robust to adversarial examples and distribution shifts when it is deployed.

The code for this baseline is available on organizers' GitHub repository\footnote{https://github.com/UrduFake/urdutask2021/blob/main/bert}.

\section{Overview of Submitted Solutions}
\label{sec:overview}

This section gives a brief overview of the systems submitted to this competition. 21 teams registered for participation, 10 teams submitted their runs for Subtask A —Abusive Language Detection, 9 teams submitted their runs for Subtask B —Threatening Language detection. Registered participants were from different countries: India, Pakistan, China, Malaysia, United Arab Emirates, Taiwan. This wide range of the regions where the interested participants were located confirms the importance of the task.
The team members came from various types of organizations: universities, research centers,
and industry.

\begin{table}[!ht]
  \caption{Approaches used by the participating systems for Subtask A: Abusive language detection}
  \resizebox{\columnwidth}{!}
  {%
   \label{tab:approaches_a}
  \begin{tabular}{lcccl}
    \toprule
System/Team Name & Feature Type & Feature Weighting Scheme & Classifying algorithm & NN-based\\
    \midrule
    
hate-alert &   laser embedding &  $N/A$   &  multi-lingual BERT   & Yes\\
SAKSHI SAKSHI &  contextual embedding & $N/A$   & Urduhack, BERT, and XLM-RoBERTa  & Yes\\
Muhammad Hamayoun & word ${1,2}$-grams  &  frequency  & SVM (sigmoid kernel) & No\\
Alt-Ed &   BoW   &  TF-IDF  &   LogReg  & No\\
Abhinav Kumar &  char ${1,6}$-grams   &  TF-IDF    & ensemble SVM+LogReg+RF   & No \\
SSNCSE\_NLP (\textit{late subm.}) & char 10-gram & TF-IDF & MLP  & Yes\\

  \bottomrule
\end{tabular}
}
\end{table}

\subsection{Approaches to Text Representation}

Participants used a variety of text representation techniques for tweet representation. Team SAKSHI SAKSHI represented tweets using contextual embedding representations that were obtained from training on an Urdu news corpus. Individual participant Muhammad Hamayoun used traditional bag-of-words representation for Subtask A and \textsc{word2vec} for  word  n-grams,  $n=1,2$, for Subtask B. The hate-alert team used pre-trained Urdu laser embeddings and multi-lingual BERT\footnote{https://huggingface.co/bert-base-multilingual-cased} pre-trained embeddings generated from an Arabic dataset. Team Alt-Ed used TF-IDF text representation. Participant Abhinav Kumar used ${1,6}$-gram character level TF-IDF features for tweet representation. A summary of approaches is presented in Tables~\ref{tab:approaches_a}, \ref{tab:approaches_b}.

\begin{table}[!h]
  \caption{Approaches used by the participating systems for Subtask B: Threatening language detection}
  \resizebox{\columnwidth}{!}
  {%
   \label{tab:approaches_b}
  \begin{tabular}{lcccl}
    \toprule
System/Team Name & Feature Type & Feature Weighting Scheme & Classifying algorithm & NN-based\\
    \midrule
    
hate-alert &   laser embedding &  $N/A$   &  multi-lingual BERT   & Yes\\
SAKSHI SAKSHI &  contextual embedding & $N/A$   &   RoBERTa   & Yes\\
Muhammad Hamayoun &  word2vec embedding  &  $N/A$ & SVM (poly d=3) & No\\
Abhinav Kumar &  char ${1,6}$-grams   &  TF-IDF    & ensemble SVM+LogReg+RF
    & No \\
SSNCSE\_NLP* & char 10-gram & TF-IDF & MLP & Yes\\

  \bottomrule
\end{tabular}
}
\end{table}

\subsection{Classification Methods}

To implement their classifiers, some participating teams used the traditional, i.e., non-neural network based machine learning algorithms, while other teams' submissions were based on various neural network architectures. 

For Subtask B, team SAKSHI SAKSHI fine-tuned a pre-trained RoBERTa model from the popular HuggingFace library\footnote{https://huggingface.co/transformers/model\_doc/roberta.html} on the Urdu news corpus
in an unsupervised manner. The same team used three transformer-based techniques for Subtask A: (i) Urduhack, (ii) BERT, and (iii) XLM-Roberta. Team hate-alert used \textsc{Hate-speech-CNERG/dehatebert-mono-arabi}\footnote{https://huggingface.co/Hate-speech-CNERG/dehatebert-mono-arabic} model which is preliminary fine-tuned on an Arabic hate speech dataset. Another participant, Muhammad Humayoun, used  SVM with sigmoid kernel for Subtask A and SVM with polynomial kernel of degree 3 for Subtask B. Participant Abhinav Kumar used an ensemble of ML models SVM + LogReg + RF for both subtasks. Similarly to one of the organizers' baseline systems, team Alt-Ed used Logistic Regression for Subtask A, which turned out to be team's best classifier for the task. 

A summary of approaches is presented in Tables~\ref{tab:approaches_a}, \ref{tab:approaches_b}, and in UrduThreat@FIRE2021 \cite{amjad2021ovv}.

\begin{table}[!h]
\caption{Final results and ranking for Subtask A: Abusive language detection}
\label{tab:overlapping2}
\begin{tabular}{@{}cccccrrrrrrrrrr@{}}
\toprule
                            & \multicolumn{3}{c}{\textit{\textbf{Private Leaderboard}}}     
                                & \multicolumn{3}{c}{\textit{\textbf{Public Leaderboard}}}                    \\
\multirow{-2}{*}{\textbf{Team Names}} & \multirow{-2}{*}{\textbf{Rank}} &
\textbf{F1 Score} & \textbf{ROC AUC} & \textbf{F1 Score} & \textbf{ROC AUC} \\ 
\hline
			hate-alert & 1 & 0.880 & 0.924 & 0.853 & 0.920 &\\
			SAKSHI SAKSHI & 2 & 0.868 & 0.935 & 0.839 & 0.934 &\\
			\textit{Org’s BERT-based sol., out-of-comp.} & $N/A$ & 0.862 & 0.921 & 0.846 & 0.923 &\\
			SATLab & 3 & 0.856 & 0.917 & 0.827 & 0.910 &\\
			Jie Yi Xiang & 4 & 0.853 & 0.914 & 0.842 & 0.915 &\\	
			Vlad Balanda & 5 & 0.849 & 0.945 & 0.824 & 0.901 &\\
			Muhammad Humayoun & 6 & 0.825 & 0.824 & 0.805 & 0.9820 &\\
			Igor Shatalin & 7 & 0.821 & 0.207 & 0.797 & 0.184 &\\
			Alt-Ed & 8 & 0.820 & 0.890 & 0.770 & 0.874 &\\
			Abhinav Kumar & 9 & 0.808 & 0.907 & 0.808 & 0.907 &\\
			ai & 10 & 0.791 & 0.908 & 0.743 & 0.875 &\\	
			\textit{Org’s LogReg v1 sol., out-of-comp.} & $N/A$ & 0.786 & 0.882 & 0.730 & 0.853 &\\
			\textit{Org’s LogReg v2 sol., out-of-comp.} & $N/A$ & 0.786 & 0.882 & 0.770 & 0.866 &\\
			SSNCSE\_NLP, \textit{late subm.}  & $N/A$ & 0.771 & 0.757 & 0.689 & 0.699 &\\
			
\bottomrule
\end{tabular}
\end{table}

  \begin{table}[!htb]
  \caption{Approaches used by the participating teams}
  \resizebox{\columnwidth}{!}
  {%
   \label{dic}
  \begin{tabular}{ccccc}
    \toprule
\textbf{System/Team Name} & \textbf{Text Representation} & \textbf{Feature Weighting Scheme} & \textbf{Classifying Algorithm} & \textbf{is NN-based?}\\
    \midrule
    
Nayel & tri-gram  &  TF-IDF  &  linear SVM with SGD & No \\
Abdullah-Khurem &  Word2Vec, GloVe, fastText &  TF-IDF     &  textCNN   & Yes \\
Hammad-Khurem & BoW  &  count \textit{(?)}   & ensemble XGBoost+LightGBM+AdaBoost    & No \\
Muhammad Homayoun  & char ${2,6}$-gram    &  \textit{N/A}   & CNN & Yes \\
Snehaan Bhawal  & transformer embeddings   &  \textit{N/A}   & MuRIL    & Yes \\
MUCIC & word- \&  char ${1,2}$-grams   & TF-IDF    & ensemble linSVM+LR+MLP+XGB+RF    & Yes (MLP) \& No \\
SOA NLP & char ${1,3}$-grams  & TF-IDF    & DNN  & Yes \\
Dinamore\&Elyasafdi\_SVC  & char 3-grams   &  TF-IDF   & SVM & No\\
MUCS & word fastText emb \& char ${2,3}$-grams  & TF-IDF for char-grams & ensemble MLP+AdaBoost+GraidentBoost+RF    & Yes (MLP) \& No \\
Iqra Ameer & transformer emb   &  \textit{N/A}   & BERT-base    & Yes \\
Sakshi Kalra & transformer emb   &  \textit{N/A}    & RoBERTa-urdu-small  & Yes \\

  \bottomrule
\end{tabular}
}
\end{table}

\begin{table}[!ht]
\caption{Final results and ranking for Subtask B: Threatening language detection}

	\label{tab:overlapping3}
\begin{tabular}{@{}cccccrrrrrrrrrr@{}}
\toprule
                            & \multicolumn{3}{c}{\textit{\textbf{Private Leaderboard}}}     
                                & \multicolumn{3}{c}{\textit{\textbf{Public Leaderboard}}}                    \\
\multirow{-2}{*}{\textbf{Team Names}} & \multirow{-2}{*}{\textbf{Rank}} &
\textbf{F1 Score} & \textbf{ROC AUC} & \textbf{F1 Score} & \textbf{ROC AUC} \\ 
\hline
			hate-alert & 1 & 0.545 & 0.810 & 0.489 & 0.798 &\\
			SATLab & 2 & 0.545 & 0.815 & 0.548 & 0.814 &\\	
			Somnath Banerjee & 3 & 0.518 & 0.791 & 0.475 & 0.776 &\\
			\textit{Org’s LogReg v2 sol., out-of-comp.} & $N/A$ & 0.514 & 0.782 & 0.451 & 0.788 &\\
			Muhammad Humayoun & 4 & 0.501 & 0.704 & 0.461 & 0.698 &\\		
			Jie Yi Xiang & 5 & 0.494 & 0.770 & 0.463 & 0.771 &\\
			SAKSHI SAKSHI & 6 & 0.492 & 0.781 & 0.534 & 0.819 &\\
			\textit{Org’s LogReg v1 sol., out-of-comp.} & $N/A$ & 0.491 & 0.769 & 0.464 & 0.795 &\\
			\textit{Org’s BERT-based sol., out-of-comp.} & $N/A$ & 0.485 & 0.700 & 0.519 & 0.765 &\\
			Vlad Balanda & 7 & 0.478 & 0.799 & 0.444 & 0.786 &\\
			Abhinav Kumar & 8 & 0.349 & 0.735 & 0.425 & 0.745 &\\	
			Owais Raza & 9 & 0.551 & 0.757 & 0.129 & 0.740 &\\	
			SSNCSE\_NLP, \textit{late subm.} & $N/A$ & 0.805 & 0.657 & 0.825 & 0.661 &\\	
\bottomrule
\end{tabular}
\end{table}

\section{Results and Discussion}
\label{sec:results}

Table {\ref{tab:overlapping2}} presents results and ranking for Abusive Language detection subask. Table {\ref{tab:overlapping3}} presents results and ranking for Threatening Language detection subtask. The systems are ranked by their F1 score on the private leaderboard.

We observe that except for one participant system, all the other participating teams’ systems outperformed the proposed LogReg baselines in terms of F1 score for Subtask A. However, only two systems, hate-alert's and SHAKSHI SHAKSHI's, outperformed the proposed BERT-based baseline.  
For Subtask B, on the contrast, quite a few systems scored below the described LogReg baseline solutions. Interestingly, even the organizers' BERT-based solution did not achieve higher scores than the LogReg baselines despite that the size of the training dataset for Subtask B was larger than the one for Subtask A.  Eventually, only the top 3 systems, hate-alert, SATLab, and participant Somnath Banerjee, achieved higher F1 scores than the organizers' \textsc{Keras}-based implementation of Logistic Regression described in Section~\ref{sec:logreg2}. Interestingly, although the two LogReg-based baselines score closely on Subtask A, their scores differ substantially for Subtask B. It might be due to the different values of the number of iterations parameter that permitted the LogReg-v2 system to converge on the larger training set in Subtask B, while in Subtask A LogReg convergence arrives sooner, partly due to a smaller training set size.

Among all the submitted runs for both sub-tasks, the hate-alert team's solution achieved the best F1 score and ranked highest. Their solutions are based on mBERT \textsc{dehatebert-mono-arabic}\footnote{https://huggingface.co/Hate-speech-CNERG/dehatebert-mono-arabic} model that is trained on an Arabic news corpus. It is plausible that the combination of a powerful deep learning model and fine-tuning on a relevant, although somewhat unexpectedly, dataset was key for the high performance. These results may open a way to further research about the effect of direct knowledge transfer among languages that use the same script, in particular, Nastal\'{i}q.

\begin{table}[!h]
\caption{Aggregated performance statistics for Subtask A: Abusive language detection. \textit{Numbers after a slash include organizers' solutions}.}
\label{tab:agg_a}
\begin{tabular}{@{}crrrr@{}}
\toprule
                                & \multicolumn{2}{c}{\textit{\textbf{Private Leaderboard}}}     
                                & \multicolumn{2}{c}{\textit{\textbf{Public Leaderboard}}}                     \\
\multirow{-2}{*}{\textbf{Agg. Stats. and Percentiles}} &
\textbf{F1 Score} & \textbf{ROC AUC} & \textbf{F1 Score} & \textbf{ROC AUC} \\ 

\hline
mean & 0.831\textit{/0.827}  & 0.830\textit{/0.844}  & 0.800\textit{/0.796} & 0.827\textit{/0.839} \\
std  & 0.033\textit{/0.035} & 0.214\textit{/0.190} & 0.049\textit{/0.049} & 0.225\textit{/0.199} \\
min  & 0.771\textit{/0.771} & 0.207\textit{/0.207} & 0.689\textit{/0.689} & 0.184\textit{/0.184} \\
10\% & 0.791\textit{/0.786} & 0.757\textit{/0.777} & 0.743\textit{/0.734} & 0.699\textit{/0.745} \\
25\% & 0.814\textit{/0.795} & 0.857\textit{/0.882} & 0.784\textit{/0.770} & 0.875\textit{/0.868} \\
50\% & 0.825\textit{/0.823} & 0.908\textit{/0.907} & 0.808\textit{/0.806} & 0.907\textit{/0.904} \\
75\% & 0.855\textit{/0.855} & 0.921\textit{/0.920} & 0.833\textit{/0.836} & 0.917\textit{/0.919} \\
90\% & 0.868\textit{/0.866} & 0.935\textit{/0.932} & 0.842\textit{/0.845} & 0.934\textit{/0.931} \\
max  & 0.880\textit{/0.880} & 0.945\textit{/0.945} & 0.853\textit{/0.853} & 0.982\textit{/0.982} \\ 
\bottomrule
\end{tabular}
\end{table}

Overall for Subtask A, 75\% of the participating systems obtained F1 score higher than 0.814 as it can be observed from the 25\textsuperscript{th} percentile in Table~\ref{tab:agg_a}. This is a good indicator that the task of abuse detection for tweets in Urdu can be achieved by automated means.  In Table~\ref{tab:overlapping2} we also observe that most of the top performing systems achieve both better F1 score and better ROC AUC for Subtask A.  

In contrast, the task of threat detection for tweets in Urdu turned out to be extremely challenging as more than 90\% of the systems could not pass the 0.8 F1 score bar as may be observed in Table~\ref{tab:agg_b}. Nevertheless, the top performing system SSNCSE\_NLP achieving F1 score of 0.805 (Table~\ref{tab:overlapping3}) provides a promising perspective that this task is also solvable with the current methods and means of NLP available for the Urdu language.  

However, at this moment it is still too soon to judge whether any of these approaches are ready to be applied ``in the wild''. While the results of over 0.88 F1 score shown by the winning hate-alert system on Subtask A are impressively high, the modest size of the provided training and testing datasets cannot guarantee the same performance on an arbitrary text input. To ensure the robustness of the presented approaches, more multifaceted research at a larger scale is needed. We see that one of the paths is a community-driven effort towards the increase of available resources and datasets in the Urdu language.

\begin{table}[]
\caption{Aggregated performance statistics for Subtask B: Threatening language detection.
\textit{Numbers after a slash include organizers' solutions}.}
\label{tab:agg_b}
\begin{tabular}{@{}crrrr@{}}
\toprule
                                & \multicolumn{2}{c}{\textit{\textbf{Private Leaderboard}}}     
                                & \multicolumn{2}{c}{\textit{\textbf{Public Leaderboard}}}                     \\
\multirow{-2}{*}{\textbf{Agg. Stats. and Percentiles}} &
\textbf{F1 Score} & \textbf{ROC AUC} & \textbf{F1 Score} & \textbf{ROC AUC} \\ 

\hline
mean & 0.528\textit{/0.521}  & 0.762\textit{/0.759}  & 0.479\textit{/0.479} & 0.761\textit{/0.766} \\
std  & 0.113\textit{/0.099} & 0.050\textit{/0.048} & 0.168\textit{/0.146} & 0.051\textit{/0.045} \\
min  & 0.349\textit{/0.349} & 0.657\textit{/0.657} & 0.129\textit{/0.129} & 0.661\textit{/0.661} \\
10\% & 0.465\textit{/0.479} & 0.699\textit{/0.701} & 0.395\textit{/0.429} & 0.694\textit{/0.706} \\
25\% & 0.492\textit{/0.491} & 0.741\textit{/0.735} & 0.448\textit{/0.451} & 0.741\textit{/0.745} \\
50\% & 0.510\textit{/0.501} & 0.776\textit{/0.770} & 0.469\textit{/0.464} & 0.774\textit{/0.776} \\
75\% & 0.545\textit{/0.545} & 0.797\textit{/0.791} & 0.523\textit{/0.519} & 0.795\textit{/0.795} \\
90\% & 0.576\textit{/0.550} & 0.810\textit{/0.808} & 0.576\textit{/0.545} & 0.815\textit{/0.811} \\
max  & 0.805\textit{/0.805} & 0.815\textit{/0.815} & 0.825\textit{/0.825} & 0.819\textit{/0.819} \\ 
\bottomrule
\end{tabular}
\end{table}

\section{Conclusion}
This paper presents a shared task in identifying threatening and abusive language in Urdu, namely, the CICLing 2021 track @ FIRE 2021 co-hosted with ODS SoC 2021.
For this track, the organizers collected two original datasets of text tweets in Urdu, one annotated for abuse (Subtask A) and the other, for threatening content (Subtask B). We also provided a training and testing split for both datasets, with the ground truth labels hidden from the participants for the testing parts of the datsets. The solutions were submitted in the form of proposed annotations for the testing sets along with the confidence score provided by the participants' systems. The submitted annotations were compared with the ground truth label to compute the F1 score, while the submitted confidence scores served for ROC AUC metric computation. The solutions were ranked by the achieved F1 scores. 

In this shared task, twenty one team from six different countries registered for the competition, and seven teams submitted their solutions. Participants used different techniques ranging from the traditional feature-crafting and application of traditional ML algorithms to word representation through pre-trained embeddings to contextual representation and end-to-end transformer based methods. An uncommon solution included an ensemble of traditional ML classifiers, SVM+LogReg+RF, whereas the particularly successful solutions used specialized BERT-based systems such as multi-lingual BERT and XLM-RoBERTa.

In the abuse detection subtask, team hate-alert outperformed all other systems with m-BERT transformer model achieving F1 score of 0.880. This and the rest of the top 3 results in Subtask A indicate that the specialized transformer based models tend to perform better compared to the feature-based traditional ML models. 

In the threat detection subtask, the hate-alert team was also a leader during the official part of the competition with the 0.545 F1 score achieved by the same m-BERT system. However, the results submitted by team SSNCSE\_NLP after the official part of the competition was closed showed a much higher F1 score of 0.805. We advert that after the end of the official part of the competition, the ground truth annotations for the testing sets were revealed to public, by this potentially putting the late submitting teams into a more advantageous position compared to the official track participants. Therefore, late submissions were not assigned a rank. 
Additionally, the technical details of SSNCSE\_NLP's solution should be enquired from the corresponding team.

This shared task aims to attract and encourage researchers working in different NLP domains to address the threatening and abusive language detection problem and help to mitigate the proliferation of offensive content on the web. Moreover, this track offers a unique opportunity to fully explore the sufficiency of textual content modality and effectiveness of fusion methods. And last but not least, the annotated datasets in Urdu are provided to the public to encourage further research and improvement of automatic detection of threatening and abusive texts in Urdu.  

\begin{acknowledgments}
This competition was organized with the support from the Mexican Government through the grant A1-S-
47854 of the CONACYT, Mexico and grants 20211784, 20211884, and 20211178 of the
Secretaría de Investigación y Posgrado of the Instituto Politécnico Nacional, Mexico.
\end{acknowledgments}

\bibliography{main}

\appendix

\end{document}